%% file: root.tex
\documentclass[letterpaper, 10 pt, conference]{ieeeconf}  % Comment this line out if you need a4paper

\IEEEoverridecommandlockouts                             
\usepackage{amsmath,amssymb,amsfonts}
\usepackage{graphicx}
\usepackage{textcomp}
\usepackage{xcolor}
\def\BibTeX{{\rm B\kern-.05em{\sc i\kern-.025em b}\kern-.08em
    T\kern-.1667em\lower.7ex\hbox{E}\kern-.125emX}}

\usepackage{comment}
\usepackage{color}
\usepackage{url}
\usepackage{booktabs}
\usepackage{acronym}
\usepackage{multirow}
\usepackage{tabularx}
\usepackage{algorithm}
\usepackage[noend]{algpseudocode}
\usepackage{graphicx}
\usepackage{subcaption}
\usepackage{xspace}
\usepackage{etoolbox,siunitx}
\usepackage{pifont}% http://ctan.org/pkg/pifont
\usepackage{wrapfig}
\usepackage{caption} 
\usepackage{adjustbox}
\usepackage{makecell}
\usepackage{afterpage}
\usepackage{mathtools}
\usepackage{stfloats}
\usepackage{svg}
\usepackage{tikz}
\usepackage{textcomp}
\usepackage{hyperref}
\usepackage{cleveref}
\usepackage{lipsum}

\usepackage{footnote}
\usepackage{fancyhdr}
\usepackage[absolute]{textpos}
\newcommand{\ourmethod}{FlexRoad\xspace}

\newcommand{\terrain}{t}
\newcommand{\road}{r}

\acrodef{SfM}[SfM]{\emph{Structure from Motion}}
\acrodef{BEV}[BEV]{\emph{Bird’s Eye View}}
\acrodef{LiDAR}[LiDAR]{\emph{Light Detection and Ranging}}
\acrodef{NURBS}[NURBS]{\emph{Non-Uniform Rational B-Splines}}
\acrodef{LoD}[LoD]{\emph{Levels of Detail}}
\acrodef{SLAM}[SLAM]{\emph{Simultaneous Localization and Mapping}}
\acrodef{MVS}[MVS]{\emph{Multi-view stereo}}
\acrodef{TSGF}[TSGF]{\emph{Two-Steps Semi-Global Filtering}}
\acrodef{Nerf}[Nerf]{\emph{Neural Radiance Fields}}
\acrodef{DEM}[DEM]{\emph{Digital Elevation Model}}
\acrodef{DTM}[DTM]{\emph{Digital Terrain Model}}
\acrodef{PMF}[PMF]{\emph{Progressive Morphological Filter}}
\acrodef{DSM}[DSM]{\emph{Digital Surface Model}}
\acrodef{DRM}[DRM]{\emph{Digital Road Model}} % new: DTM + bridges
\acrodef{PMHR}[PMHR]{\emph{Improved Progressive Morphological Filter based on Hierarchical Radial Basis Function Interpolation}}
\acrodef{DOP}[DOP]{\emph{Digital Orthophoto}}
\acrodef{TIN}[TIN]{\emph{Triangulated Irregular Network}}
\acrodef{RGT}[RGT]{\emph{Regular Grid Trianglation}}
\acrodef{MLP}[MLP]{\emph{Multi-Layer Perceptron}}
\acrodef{RMSE}[RMSE]{\emph{Root Mean Squared Error}}
\acrodef{MAE}[MAE]{\emph{Mean Absolute Error}}
\acrodef{MAD}[MAD]{\emph{Mean Absolute Deviation}}
\acrodef{SMRF}[SMRF]{\emph{Simple Morphological Filter}}
\acrodef{SSIM}[SSIM]{\emph{Structural Similarity Index Measure}}
\acrodef{FILTER}[ECSRC]{\emph{Elevation-Constrained Spatial Road Clustering}}
\acrodef{RANSAC}[RANSAC]{\emph{Random Sample Consensus}}
\acrodef{DSC-SIFI}[SIFI]{Fabulous Sindelfingen}
\acrodef{DSC-STR}[STR]{Stunning Stuttgart}
\acrodef{DSC-MUC}[MUC]{Great Munich}
\acrodef{DSC-BER}[BER]{Vibrant Berlin}
\acrodef{RTK}[RTK]{Real-Time Kinematic Positioning}
\acrodef{GNSS}[GNSS]{Global Navigation Satellite System}
\acrodef{IMU}[IMU]{Inertial Measurement Unit}
\acrodef{GCP}[GCP]{Ground Control Point}
\acrodef{UAV}[UAV]{Unmanned Aerial Vehicle}
\acrodef{DSC}[DSC3D]{DeepScenario Open 3D Dataset}
\acrodef{GeoRoad}[GeRoD]{GeoRoad Dataset}

\newcolumntype{P}[1]{>{\centering\arraybackslash}p{#1}}

\begin{document}

\title{Shape Your Ground: Refining Road Surfaces Beyond Planar Representations
}

\author{\IEEEauthorblockN{Oussema Dhaouadi$^{1,2,3}$}
\and
\IEEEauthorblockN{Johannes Meier$^{1,2,3}$}
\and
\IEEEauthorblockN{Jacques Kaiser$^{1}$}
\and
\IEEEauthorblockN{Daniel Cremers$^{2,3}$}
}

\author{Oussema Dhaouadi$^{1,2,3,\dagger}$, Johannes Meier$^{1,2,3}$, Jacques Kaiser$^{1}$, Daniel Cremers$^{2,3}$ \\
$^{1}$ DeepScenario\quad $^{2}$ TU Munich\quad $^{3}$ Munich Center for Machine Learning \\ \\ {\small Project Page:} {\footnotesize \textcolor[HTML]{F1238F}{\textbf{\url{https://deepscenario.github.io/FlexRoad/}}}}
\thanks{$\dagger$ Corresponding author.}%
\thanks{TUM: {\tt \{oussema.dhaouadi, j.meier, cremers\}@tum.de}}
\thanks{DeepScenario: \tt firstname@deepscenario.com}
}

\maketitle
\begin{textblock}{175}(16,263)   % {width}(x,y) coordinates
\scriptsize
\setlength{\fboxsep}{3pt}
\framebox{\parbox{\textwidth}{
Accepted at IEEE Intelligent Vehicles Symposium (IV) 2025. \copyright 2025 IEEE.  Personal use of this material is permitted.  Permission from IEEE must be obtained for all other uses, in any current or future media, including reprinting/republishing this material for advertising or promotional purposes, creating new collective works, for resale or redistribution to servers or lists, or reuse of any copyrighted component of this work in other works.”
}}
\end{textblock}

\input{chapters/0_abstract}
\input{chapters/1_introduction}
\input{chapters/2_related_work}
\input{chapters/3_method}
\input{chapters/4_experiments}
\input{chapters/5_conclusion}

\bibliographystyle{IEEEtran}
\bibliography{references}

\end{document}

%% file: chapters/0_abstract.tex
\begin{abstract}
Road surface reconstruction from aerial images is fundamental for autonomous driving, urban planning, and virtual simulation, where smoothness, compactness, and accuracy are critical quality factors. Existing reconstruction methods often produce artifacts and inconsistencies that limit usability, while downstream tasks have a tendency to represent roads as planes for simplicity but at the cost of accuracy. We introduce \ourmethod, the first framework to directly address road surface smoothing by fitting \acf{NURBS} surfaces to 3D road points obtained from photogrammetric reconstructions or geodata providers. Our method at its core utilizes the \acf{FILTER} algorithm for robust anomaly correction, significantly reducing surface roughness and fitting errors. To facilitate quantitative comparison between road surface reconstruction methods, we present \acf{GeoRoad}, a diverse collection of road surface and terrain profiles derived from openly accessible geodata. Experiments on \ac{GeoRoad} and the photogrammetry-based \acf{DSC} demonstrate that \ourmethod considerably surpasses commonly used road surface representations across various metrics while being insensitive to various input sources, terrains, and noise types. By performing ablation studies, we identify the key role of each component towards high-quality reconstruction performance, making \ourmethod a generic method for realistic road surface modeling.
\end{abstract}

%% file: chapters/1_introduction.tex
\section{Introduction}

\begin{figure}[t]
    \centering
    \begin{subfigure}[b]{0.48\linewidth}
        \centering
        \includegraphics[width=\textwidth]{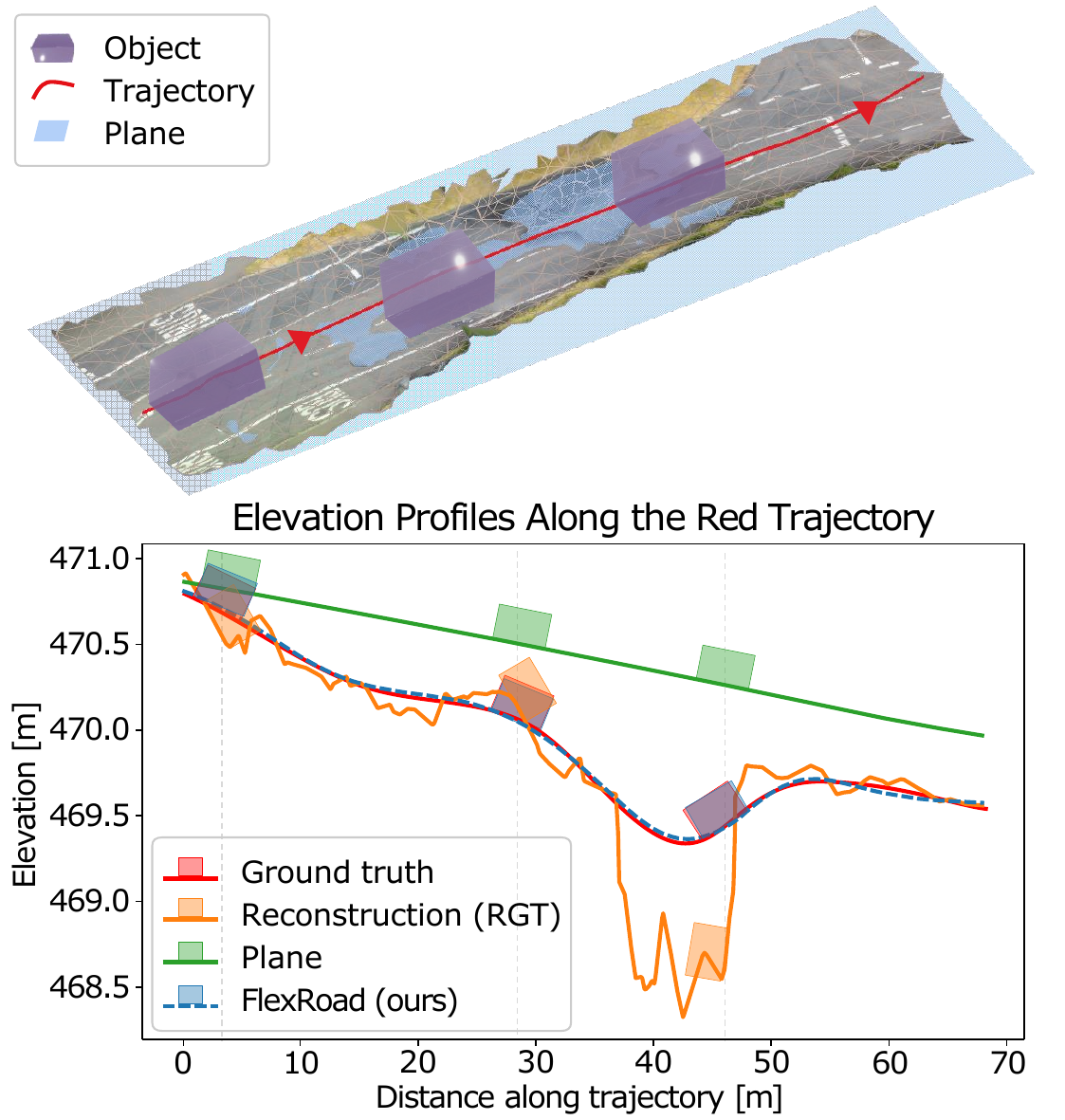}
        \caption{Steep elevation profiles in \acs{DSC} \cite{dsc3d} (Stunning Stuttgart:~STR)}
        \label{fig:unsmooth_road_stuttgart}
    \end{subfigure}
    \hfill
    \begin{subfigure}[b]{0.48\linewidth}
        \centering
        \includegraphics[width=\textwidth]{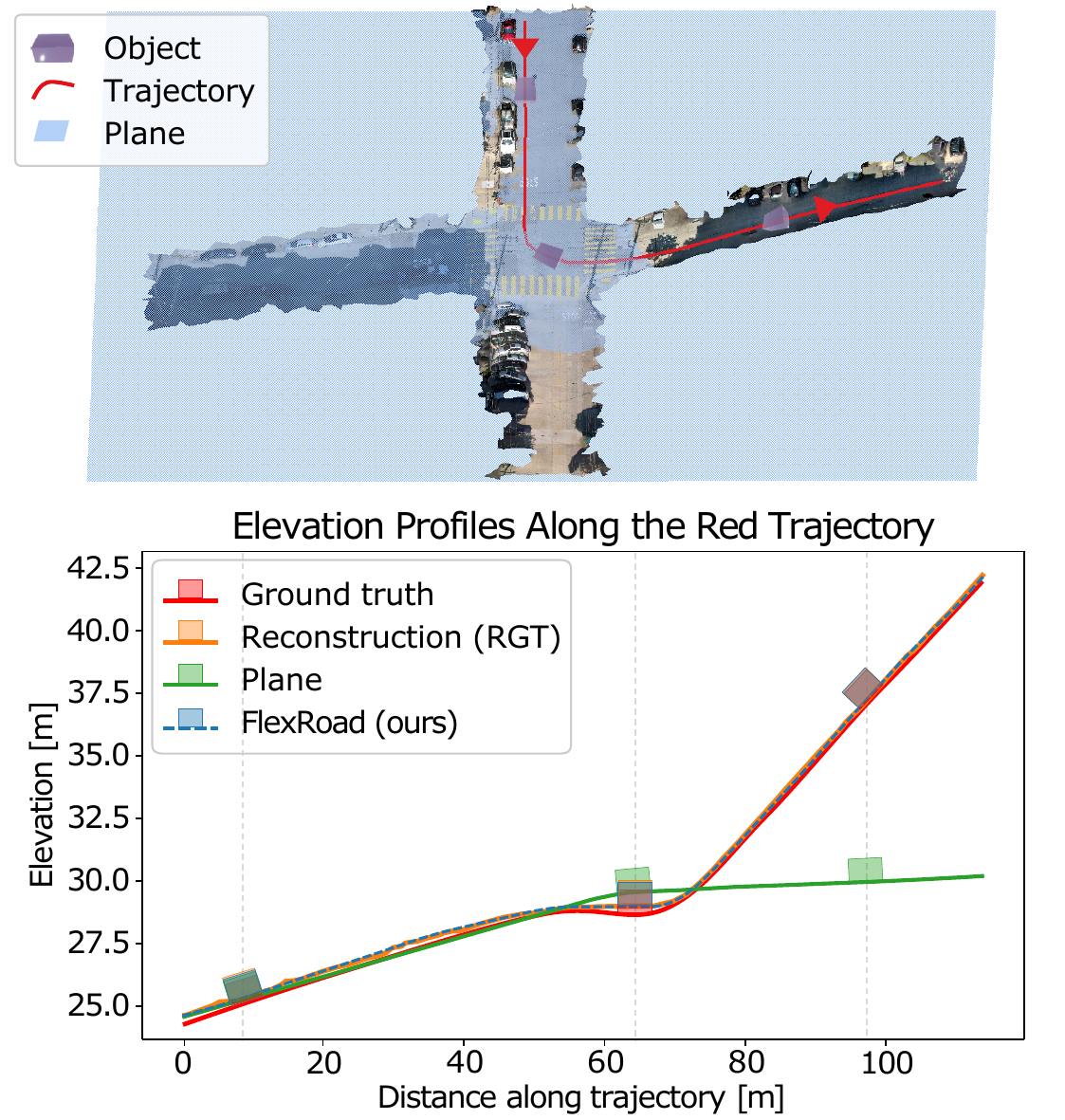}
    \caption{Surface discontinuities in \acs{DSC} \cite{dsc3d} (Visionary San Francisco:~SFO)}
        \label{fig:steep_road_sfo}
    \end{subfigure}
   \caption{\textbf{Comparison of road profiles} along the vehicle trajectory (red): raw reconstruction (orange), planar approximation (green), and \ourmethod (dashed blue). \acs{RGT} denotes a direct meshing method from \acs{SfM} outputs, while the Plane method fits a simple plane to the reconstructed geometry.}
    \label{fig:challenging_cases}
    \vspace{0.5\baselineskip}
\end{figure}

Road surface reconstruction from airborne imagery is one of the primary urban digitization challenges and is crucial in autonomous navigation, infrastructure inspection, and city planning \cite{road_2023}. While modeling the fine road geometry with cracks or depressions from aerial perspectives is challenging, most downstream applications primarily require smooth surface representations. For instance, 3D object detection for autonomous driving often needs to be refined by projecting onto road surfaces~\cite{dsc3d} (\Cref{fig:challenging_cases}). These estimates employ fixed surface normals to correct the position and orientation of objects detected, and therefore surface smoothness plays a crucial role in ensuring stable tracking and trajectory estimation.

State-of-the-art reconstruction pipelines such as COLMAP~\cite{colmap_2016} and OpenSfM~\cite{opensfm_2020} yield road surfaces with artifacts like bumps and holes (\Cref{fig:unsmooth_road_stuttgart}). The causes are threefold (not mutually exclusive): elevation mixing when downsampling, interference from off-road objects, and sparse aerial coverage reconstruction artifacts. Classic planar approximations are commonly employed in downstream operations like city modelings \cite{nyc_citygml_2017} to ensure smoothness at the expense of important geometric variations, particularly over areas of extreme elevaion differences (\Cref{fig:steep_road_sfo}). Recent neural approaches like NeRF \cite{nerf_2021} and Gaussian Splatting \cite{gaussian_splatting_2023}, while enhancing novel view synthesis, are not geometrically accurate in sparsely viewed areas.

We address these challenges using a refinement procedure that optimizes both surface reconstruction quality and geometric smoothness simultaneously. Our method is particularly well-suited to 3D reconstruction of road surfaces that produces smooth transitions to surrounding terrain. Rather than aiming at fine-grained feature reconstruction from high-altitude imagery, our method prioritizes geometric consistency while preserving essential surface characteristics.

The contributions of this paper are threefold:

\begin{itemize}
    \item We present \ourmethod, the first pipeline specifically designed to refine road surfaces using \ac{NURBS} fitting. Our method combines geometric filtering for noise removal, coarse feature preservation, and adaptive sampling for optimizing mesh complexity.

    \item We introduce \acf{FILTER}, a novel algorithm that effectively removes noisy road classifications including points from facades, trees, parked vehicles, and other non-road elements.

    \item We present \ac{GeoRoad}, a curated benchmark dataset including aerial imagery, elevation models derived from \ac{LiDAR} scans, and ground truth road 3D points. We show the improved performance of our method across different data sources on both \ac{GeoRoad} and the photogrammetry-based \ac{DSC} \cite{dsc3d} dataset.

\end{itemize} 

%% file: chapters/2_related_work.tex
\section{Related Work}
\label{sec:related_work}

Reconstruction of road surfaces is crucial for various applications, ranging from autonomous navigation to infrastructure inspection and smart city solution development. Although conventional techniques have been promising, they usually suffer from serious challenges in the context of road surfaces, such as textureless regions and repetitive patterns, resulting in poor or deformed reconstructions. In this section, we present techniques closely related to our method and corresponding techniques in similar domains such as vehicle-based reconstruction and terrain modeling.

\subsection{3D Road Surface Reconstruction}

Recent advances in road surface reconstruction have leveraged multi-modality and multi-representation data. Neural implicit representations have been particularly promising, with methods like RoMe~\cite{rome_2023} combining RGB and semantic data to handle dynamic occlusions and lighting variations. EMIE-MAP~\cite{emie-map_2024} uses a hybrid of explicit meshes and implicit encoding and shows robust performance in urban environments through height-residual learning. Later, RoGS~\cite{rogs_2024} introduced 2D Gaussian surfels to model roads, parameterizing position, color, scale, and semantics and adding \ac{LiDAR} supervision for height.

Yet, these methods are mostly designed for ground-level, vehicle-mounted collections with the premise of high scene overlap and forward-facing cameras. Consequently, they are not as well-suited for aerial-based collections with varying issues such as high altitudes, sparse coverage, and oblique views. Our system specifically overcomes these shortcomings by refining noisy reconstructions, producing complete and smooth road manifolds with geometric correctness and semantic coherence.

\subsection{Digital Elevation Modeling}

\acfp{DEM} are raw data representations in geospatial analysis, where \acfp{DSM} captures all surface features and \acfp{DTM} captures ground topology only~\cite{sparsity_driven_dtm_2018}. Traditional DSM-to-DTM conversion is based on height analysis~\cite{vosselman_2000} or morphological filtering~\cite{pmf_2003}, while deep learning methods~\cite{dl_dtm_2016,als2dtm_2022} achieve sub-metric accuracy through point cloud classification.

These methods employ two primary terrain representations: regular grids through \ac{RGT}, whose cells are of equal size divided into triangles, and \acfp{TIN} that allow adaptively sized triangles according to terrain complexity. Regular grids are computationally efficient at the expense of accuracy in areas of high elevation variation, while \acp{TIN}~\cite{delaunay_1934} follow terrain variations more accurately at higher computational cost. There are practical limitations as well: downsampling through cell-grid aggregation introduces artifacts when cells contain both road and non-road points (due to buildings, trees, guard rails, or parked vehicles), and \acp{DSM} include non-road objects that compromise geometry accuracy.

Our method can leverage these elevation models as initial reconstruction while providing a refined surface representation that eliminates this type of structural noise and enforces smoothness and geometric coherence. Moreover, the parametric representation of the refined surface using our method enables multi-resolution meshing through computationally low-cost forward passes. Our adaptive sampling method yields a hybrid representation of high-resolution road surfaces and low-resolution terrain through two forward passes with a high reduction in computation overhead.

\subsection{Parametric Road Surface Representation}

Parametric surface modeling using B-splines has been explored for road surface modeling in vehicle-centric applications. Early work~\cite{bspline_road_2008,bspline_road_2009} has demonstrated B-spline fitting to stereo-derived point clouds for non-planar road modeling, with an emphasis on local surface models from car-mounted sensors. While these methods generate good results for vehicle-view models, they suffer from several limitations: they rely on stereo camera configurations for depth estimation, are limited to local road sections, and are not formulated to handle the challenges of aerial views.

\acf{NURBS}~\cite{nurbs_book_1997} are a more general class of parametric surfaces than B-splines, with increased flexibility provided by control point manipulation and weighting. Although common in computer-aided design and terrain modeling~\cite{nurbs_integrated_fitting_fairing_2018}, \ac{NURBS} have not previously been used to model road surfaces from aerial imagery. This is the first study to utilize \ac{NURBS} to refine road surface modeling from aerial data, integrating orthophoto-based semantic segmentation and elevation data. This novel application of \ac{NURBS}, motivated by the natural smoothness of road infrastructure, enables us to represent entire road networks as a single low-weight surface and represent complex geometries in large environments.

\subsection{Ground Surface Assumptions}

A majority of computer vision tasks assume simplified models for the ground surface, typically piecewise-planar or planar surfaces. Recent monocular 3D object detection efforts~\cite{monoground_2022,mvdet_2023,ground_plane_matters_2022} show the utility of using ground plane assumptions. While methods like~\cite{noise_resilient_piecewise_plane_2013} extend the same to piecewise linear functions using RANSAC-based estimation, such approximations fail to model real-world road geometry with high accuracy, particularly in challenging cases with extreme elevation changes. Our method overcomes such limitations by providing a more compliant surface model that can better accommodate terrain variations with ease. We believe that \ourmethod has the potential to enhance the robustness of existing computer vision systems, especially for steep roads where the conventional planar assumptions cannot handle the geometric complexity of the surface.

%% file: chapters/3_method.tex
\section{Method} \label{sec:method}
\begin{figure*}[!tb]
\centering
\includegraphics[width=1\linewidth]{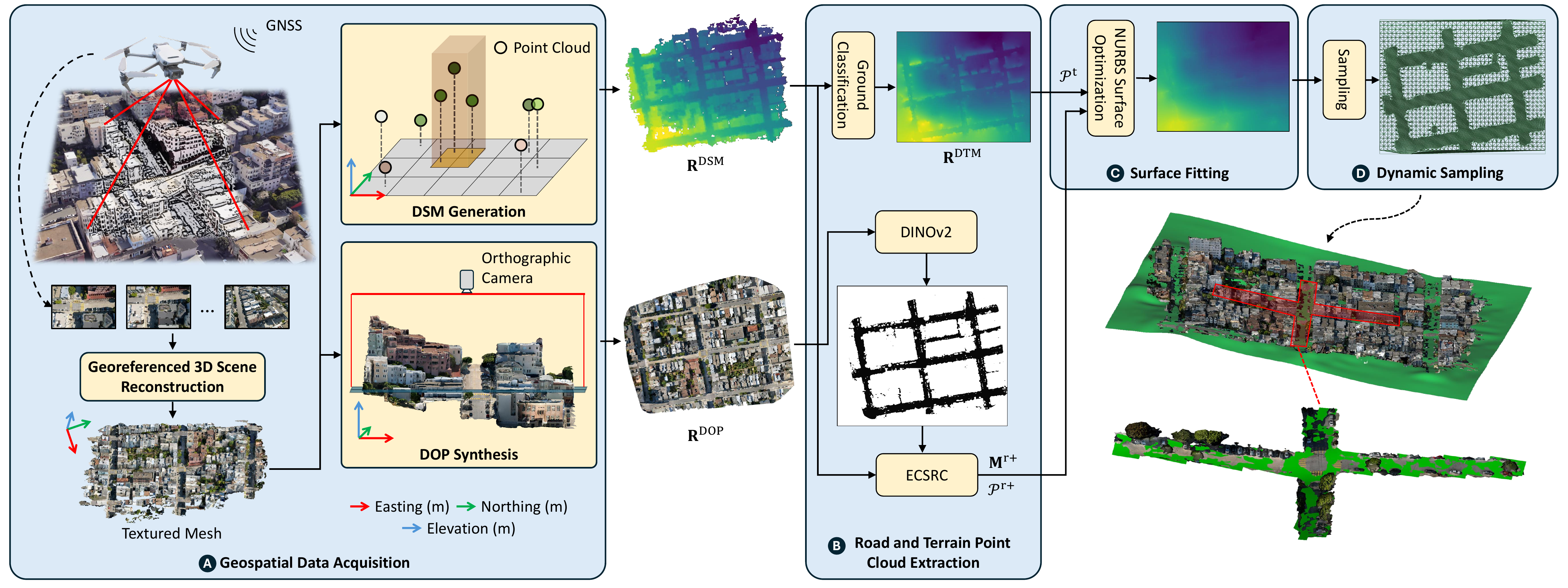}
\caption{\textbf{Overview of our pipeline.} Our workflow has four main steps: (A) Geospatial data acquisition either through 3D scene reconstruction from \acs{UAV} images (yielding a textured mesh, \acs{DOP}, and \acs{DSM}) or directly from geodata; (B) road and terrain point cloud extraction through DINOv2~\cite{dinov2_2023} and \acs{FILTER}-based masking; (C) surface fitting through \acs{NURBS} representation; and (D) dynamic surface sampling for efficient mesh generation.}

\label{fig:pipeline}
\vspace*{-1\baselineskip}
\end{figure*}

Our pipeline consists of four main sequential stages, progressing from initial data acquisition to the final road mesh generation, as illustrated in \Cref{fig:pipeline}.

\subsection{Notations}
Throughout the paper, we denote $\bullet|_{xy}$ as the projection operator onto the XY plane, mapping a 3D point to its x and y components as a 2D vector, and $\bullet|_{z}$ as the operator extracting the scalar z-coordinate. Specifically, for a 3D point at grid cell position $(i, j)$, we write $\mathbf{p}_{ij} = [\mathbf{p}_{ij}|_{x}, \mathbf{p}_{ij}|_{y}, \mathbf{p}_{ij}|_{z}]^\top \in \mathbb{R}^3$ with its XY projection as $\mathbf{p}_{ij}|_{xy} = [\mathbf{p}_{ij}|_{x}, \mathbf{p}_{ij}|_{y}]^\top$ and its elevation as $\mathbf{p}_{ij}|_{z}$. 

\subsection{Geospatial Data Acquisition} \label{sec:input}
Our surface reconstruction pipeline operates on two fundamental inputs: \ac{DSM} and \ac{DOP}. These inputs can be obtained either through existing geodata portals or generated via aerial photogrammetry. In the latter approach, we process \ac{GNSS}-tagged \ac{UAV} imagery through a sequential pipeline of \ac{SfM}~\cite{schonberger2016structure} for sparse reconstruction and camera pose estimation, followed by \ac{MVS}~\cite{openmvs_2020}. The resulting point cloud undergoes Poisson surface reconstruction~\cite{poisson_2013} to generate the final mesh representation. Robust positioning is achieved by employing both direct (RTK-\ac{GNSS}, \ac{IMU}) and indirect (\acp{GCP}) georeferencing techniques.
We acquire a dense point cloud $\mathcal{P}$ through uniform sampling of the textured mesh. We perform two parallel processing operations on the point cloud: rasterization into a \ac{DSM} matrix $\mathbf{R}^{DSM} \in \mathbb{R}^{W \times H}$ of cell size $s_w^{DSM} \times s_h^{DSM}$, with each cell elevation computed from the maximum height value of enclosed points, and orthographic projection to generate a georeferenced \ac{DOP}. Orthographic representation has spatial fidelity guaranteed by removing perspective distortions, thereby guaranteeing constant scale throughout the entire surface representation.

\subsection{Road and Terrain Point Cloud Extraction} \label{sec:road_ground_extraction}
With DINOv2 \cite{dinov2_2023}, we segment the road from the \ac{DOP} to obtain a binary road mask $\mathbf{M}^{\road}$, where 1 represents road points and 0 represents terrain area. To align spatial dimensions, we re-sample the road mask $\mathbf{M}^{\road}$ to match the \ac{DSM} resolution $\mathbf{R}^{DSM}$ using nearest-neighbor interpolation. The road point set $\mathcal{P}^{\road}$ is defined as:
\begin{adjustbox}{width=1.0\linewidth}
\begin{minipage}{1.0\linewidth}
\vspace{0.5\baselineskip}
\begin{align}
    \mathcal{P}^{\road} = \Bigg\{ \mathbf{p}_{ij} : 
    \begin{array}{l}
    \mathbf{p}_{ij}|_{x} = iW, \, \mathbf{p}_{ij}|_{y} = jH, \, \\
    \mathbf{p}_{ij}|_{z} = \mathbf{R}^{DSM}_{ij}, \, \mathbf{M}^{\road}_{ij} = 1
    \end{array} \Bigg\} \,.
\end{align}
\vspace{0.5\baselineskip}
\end{minipage}
\end{adjustbox}

We correct segmentation artifacts and off-road objects (e.g., facades, trees, cars) using \acf{FILTER}, our noise removal algorithm that fuses region-growing concepts with elevation-based clustering, as described in \Cref{alg:FILTER}. The combined strategy leverages spatial connectivity and elevation consistency to effectively detect and remove noise while preserving the road structure.

\begin{algorithm}[ht]
\caption{\ac{FILTER} Algorithm}
\label{alg:FILTER}
\begin{algorithmic}[1]
\State \textbf{Input:} $\mathcal{P}^{\road}$: road point cloud; $\theta_{xy}$: maximum neighbor distance in the XY plane; $\theta_z$: maximum neighbor distance along the Z-axis
\State \textbf{Output:} labels $\mathbf{L}$
\State $\mathcal{N} \gets \texttt{getNeighbors}(\mathcal{P}^{\road}, \theta_z)$ 
\State initialize $\mathbf{L} = \mathbf{0}$; $\ell \gets 1$ \Comment{{}\parbox[t]{.37\linewidth}{\footnotesize Initialize labels of same size as $\mathcal{P}^{\road}$ to 0 (non-road)}}
\For{$\mathbf{p}_{ij} \in \mathcal{P}^{\road}$}
    \If{$\mathbf{L}_{ij} = 0$}
        \State $\mathcal{V} \gets \text{set}()$ \Comment{{}\parbox[t]{.37\linewidth}{\footnotesize $\mathcal{V}$ set of visited points}}
        \State $\mathcal{Q} \gets \text{set}()$ \Comment{{}\parbox[t]{.37\linewidth}{\footnotesize $\mathcal{Q}$ set of points to visit}}
        \State $\mathcal{Q}.\text{add}(\mathbf{p}_{ij})$
        \While{$\mathcal{Q} \neq \varnothing$}
            \State $\mathbf{p}_{i^{*}j^{*}} \gets \mathcal{Q}.\text{pop}()$ 
            \If{$\mathbf{p}_{i^{*}j^{*}} \notin \mathcal{V}$}  
                \State $\mathcal{V}.\text{add}(\mathbf{p}_{i^{*}j^{*}})$
                \State $\mathbf{L}_{i^{*}j^{*}} \gets \ell$ 
                \State $\mathcal{Q} \gets \mathcal{Q} \cup \mathcal{N}[\mathbf{p}_{i^{*}j^{*}}]$
            \EndIf
        \EndWhile
        \State $\ell \gets \ell + 1$ 
    \EndIf
\EndFor
\State $\mathbf{L} \gets \texttt{mergeClusters}(\mathcal{P}^\road, \mathbf{L}, \theta_{xy}, \theta_{z})$ 
\State $\mathbf{L} \gets \texttt{cleanClusters}(\mathcal{P}^\road, \mathbf{L})$
\end{algorithmic}
\end{algorithm}
For each point $\mathbf{p}_{ij} \in \mathcal{P}^{\road}$, the function \texttt{getNeighbors} considers only direct adjacent grid neighbors $A(i,j)$ and their elevation differences. Specifically, it computes the neighborhood set:
\begin{adjustbox}{width=1\linewidth}
\begin{minipage}{1.0\linewidth}
\begin{equation}
    \mathcal{N}[\mathbf{p}_{ij}] = \Bigg\{\mathbf{q}_{i'j'} \in \mathcal{P}^{\road} : \begin{array}{l}
    |\mathbf{q}_{i'j'}|_{z} - \mathbf{p}_{ij}|_{z}| \leq \theta_z, \\
    (i',j') \in A(i,j)
    \end{array} \Bigg\} \, ,
    \label{eq:neighbors}
\end{equation}
\end{minipage}
\end{adjustbox}

\begin{adjustbox}{width=1\linewidth}
\begin{minipage}{1.04\linewidth}
\begin{align}
    A(i,j) = \{i-1, i, i+1\} \times \{j-1, j, j+1\} \setminus \{(i, j)\} \, ,
\end{align}
\vspace{0.0\baselineskip}
\end{minipage}
\end{adjustbox}
where $\theta_z$ represents the elevation threshold.

Subsequently, the \texttt{mergeClusters} function introduces an additional spatial proximity criterion:
\begin{equation}
    \| \mathbf{q}_{ij}|_{xy} - \mathbf{p}_{ij}|_{xy} \|_2 \leq \theta_{xy} \, ,
\end{equation}
where $\theta_{xy}$ defines the maximum allowable planar distance between points from different clusters. This spatial condition, combined with the elevation criterion, guides the merging of initially formed clusters into larger, coherent groups.

The \texttt{cleanClusters} function then filters the results by eliminating small clusters and preserving only the top-$k$ largest ones, yielding the filtered road point cloud $\mathcal{P}^{\road+}$ with mask $\mathbf{M}^{\road+}$.

The terrain point cloud $\mathcal{P}^{\terrain}$, derived from Vosselman et al.'s~\cite{vosselman_2000} ground classification algorithm, serves as support during surface fitting and is obtained from the \ac{DTM} raster $\mathbf{R}^{DTM}$ of size $s_w^{DTM}\times s_h^{DTM}$:

\begin{adjustbox}{width=1\linewidth}
\begin{minipage}{1.0\linewidth}
\vspace{0.5\baselineskip}
\begin{equation}
    \mathcal{P}^{\terrain} = \Bigg\{ 
    \mathbf{p}_{ij} : 
    \begin{array}{l}
    \mathbf{p}_{ij}|_{x} = iW, \, \mathbf{p}_{ij}|_{y} = jH, \, \\
    \mathbf{p}_{ij}|_{z} = \mathbf{R}^{DTM}_{ij}, \, \mathbf{M}^{\road +}_{ij} = 0
    \end{array} \Bigg\} \,.
\end{equation}
\vspace{-0.5\baselineskip}
\end{minipage}
\end{adjustbox}

\subsection{Surface Fitting}
We model the road surface using a \ac{NURBS} \cite{nurbs_book_1997} surface because of its ability to represent smooth surfaces from noisy and scattered point clouds. The surface is represented using polynomial basis functions (B-splines) $N_i^p$ and $N_j^q$ of degree $p$ and $q$ over parametric domain $(u,v)$. The surface is defined by knot vectors $\mbox{$\mathbf{U} = [u_1,., u_{n+p+2}]$}$, $\mbox{$\mathbf{V} = [v_1,., v_{m+q+2}]$}$, control points matrix $\mathbf{P} \in \mathbb{R}^{n \times m \times 3}$ and weights matrix $\mathbf{W} \in \mathbb{R}^{n\times m}$. Here, $\mathbf{P}_{ij} \in \mathbb{R}^3$ represents the individual control point at indices $(i,j)$ in matrix $\mathbf{P}$, and $w_{ij}$ denotes the corresponding scalar weight at indices $(i,j)$ in matrix $\mathbf{W}$. The surface can be expressed as~\cite{nurbs_book_1997}:
\begin{align}
    S_{\mathbf{\Theta}}(u, v)=\frac{\sum_{i=0}^n \sum_{j=0}^m N_i^p(u) N_j^q(v) w_{i j} \mathbf{P}_{i j}}{\sum_{i=0}^n \sum_{j=0}^m N_i^p(u) N_j^q(v) w_{i j}} \label{eq:nurbs}\, .
\end{align}
The parameter set $\mathbf{\Theta}$ includes $\mathbf{U}$, $\mathbf{V}$, $\mathbf{W}$, and $\mathbf{P}$. The basis functions are computed using Cox-de Boor recursion~\cite{nurbs_book_1997}:
\begin{adjustbox}{width=1\linewidth}
\begin{minipage}{1.05\linewidth}
\begin{align}
    \begin{gathered}
    N_i^p(u)=\frac{u-u_i}{u_{i+p}-u_i} N_i^{p-1}(u)+\frac{u_{i+p+1}-u}{u_{i+p+1}-u_{i+1}} N_{i+1}^{p-1}(u) \\
    N_i^0(u)=\begin{cases}
            1 & \text{if }u_i\leq u\leq u_{i+1}\\ 
            0 & \text{otherwise}
        \end{cases} \, .
    \end{gathered} \label{eq:cox_de_boor}
\end{align}
\vspace{0.0\baselineskip}
\end{minipage}
\end{adjustbox}
We fit a differentiable \ac{NURBS} surface \cite{nurbs_diff_2022} using gradient descent optimization with a composite loss function:
\begin{align}
    \mathcal{L}(\mathbf{\Theta}) &= \mathcal{L}^{\road}(\mathbf{\Theta}) + \lambda_{\terrain} \mathcal{L}^{\terrain}(\mathbf{\Theta}) + \lambda_{reg} \mathcal{L}^{reg}(\mathbf{\Theta})\, ,
\end{align}
where $\mathcal{L}^{\road}$ minimizes road point discrepancy using $L_1$ loss for outlier robustness:
\begin{align}
   \mathcal{L}^{\road}(\mathbf{\Theta}) &= \frac{1}{HW}\sum_{i,j}^{HW} \lvert \mathbf{M}_{ij}^{\road +} (\mathbf{R}^{DSM}_{ij} - \mathbf{R}^{S}_{ij} (\mathbf{\Theta})) \rvert \, ,
\end{align}
$\mathcal{L}^{\terrain}$ ensures smooth and seamless road-terrain transitions:
\begin{adjustbox}{width=1\linewidth}
\begin{minipage}{1.01\linewidth}
\begin{align}
   \mathcal{L}^{\terrain}(\mathbf{\Theta}) &= \frac{1}{HW}\sum_{i,j}^{HW} \lvert (1 - \mathbf{M}_{ij}^{\road +}) (\mathbf{R}^{DTM}_{ij} - \mathbf{R}^{S}_{i, j}(\mathbf{\Theta})) \rvert \, , \label{eq:loss_terrain}
\end{align}
\vspace{0.0\baselineskip}
\end{minipage}
\end{adjustbox}
and $\mathcal{L}^{reg}$ prevents excessive elevation differences between adjacent control points:
\begin{adjustbox}{width=1\linewidth}
\begin{minipage}{1.3\linewidth}
\begin{align}
   \mathcal{L}^{reg}(\mathbf{\Theta}) &= \frac{1}{HW}\sum_{i,j}^{HW} \left[ \max_{l \in A(i,j)} (\mathbf{P}_{ij} - \mathbf{P}_{l})|_{z} - \min_{l \in A(i,j)} (\mathbf{P}_{ij} - \mathbf{P}_{l})|_{z} \right] ^2  \label{eq:loss_ctrl} \, .
\end{align}
\vspace{0.0\baselineskip}
\end{minipage}
\end{adjustbox}
\noindent Here, we define $\mathbf{R}^{S}_{i, j}(\mathbf{\Theta})$ as the rasterized fitted \ac{NURBS} surface with dimensions matching those of the \ac{DSM} and \ac{DTM}.

\subsection{Dynamic Surface Sampling}
Dynamic sampling uses high-resolution sampling rate $s^{\road}$ on road and low-resolution sampling rate $s^{\terrain}$ on terrain to create rasters $\mathbf{R}^{S, \road}(\mathbf{\Theta})$ and $\mathbf{R}^{S, \terrain}(\mathbf{\Theta})$. They are combined to produce a \ac{TIN} using the road mask:
\begin{adjustbox}{width=1\linewidth}
\begin{minipage}{1.11\linewidth}
\begin{align}
    \{\mathbf{p}_{ij} \in \mathbf{R}^{S, \road}(\mathbf{\Theta}) | \mathbf{M}^{\road +}_{ij} = 1\} \cup \{\mathbf{p}_{ij} \in \mathbf{R}^{S, \terrain}(\mathbf{\Theta}) | \mathbf{M}^{\road +}_{ij} = 0 \} \, ,
\end{align}
\vspace{0.0\baselineskip}
\end{minipage}
\end{adjustbox}
followed by Delaunay triangulation~\cite{delaunay_1934} to build the final mesh.

%% file: chapters/4_experiments.tex
\section{Experimental Evaluation} \label{sec:results}
We present a comprehensive evaluation of \ourmethod through experiments on two datasets, examining its performance against two baselines through quantitative metrics and qualitative assessments.

\subsection{Datasets}

\subsubsection{\acf{GeoRoad}}
\begin{figure}[!ht]
\vspace{0.5\baselineskip}
    \centering
    \begin{subfigure}[b]{0.24\linewidth}
        \centering
        \fbox{\includegraphics[width=\textwidth]{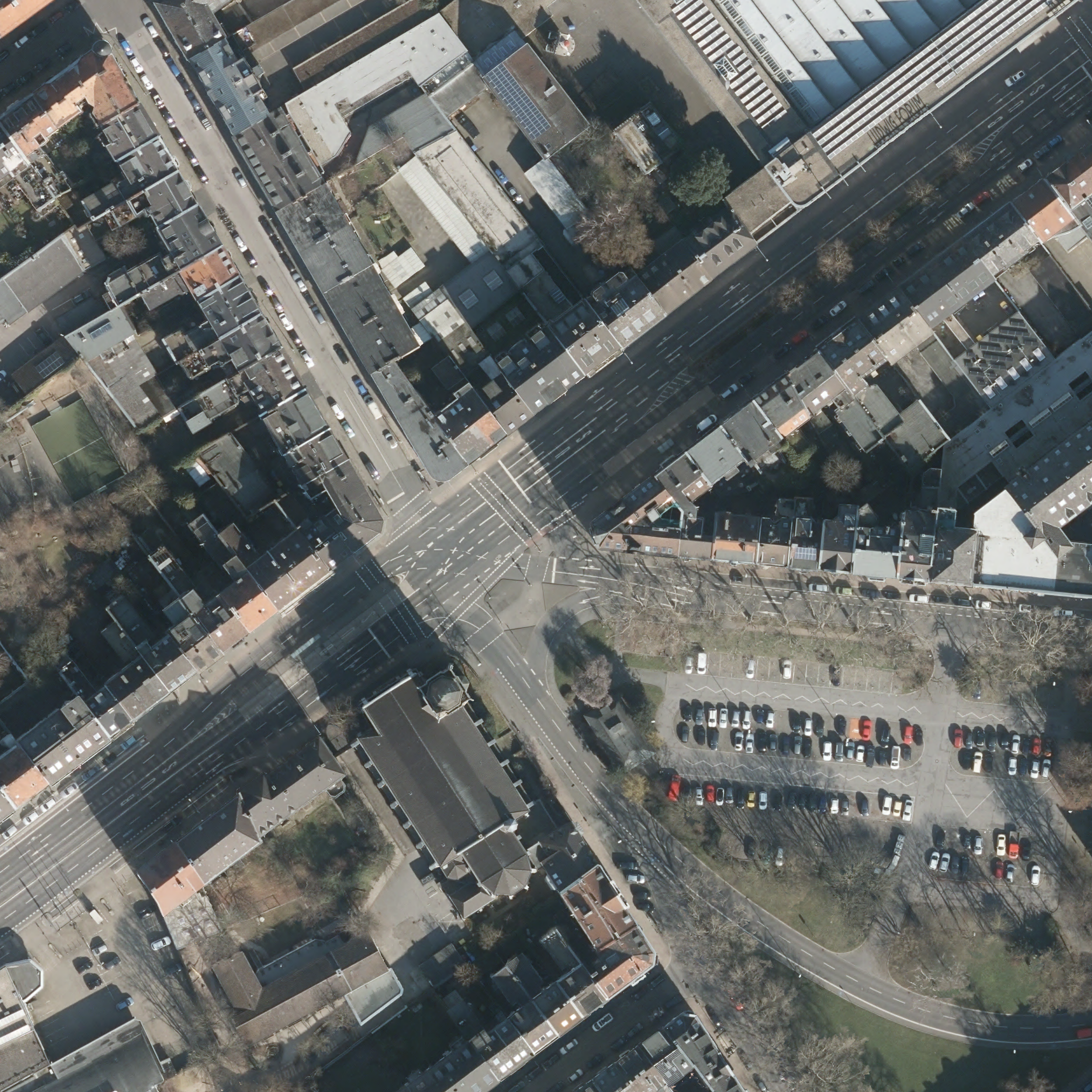}}
        \caption{\ac{DOP}}
        \label{fig:sub1}
    \end{subfigure}
    \begin{subfigure}[b]{0.24\linewidth}
        \centering
        \fbox{\includegraphics[width=\textwidth]{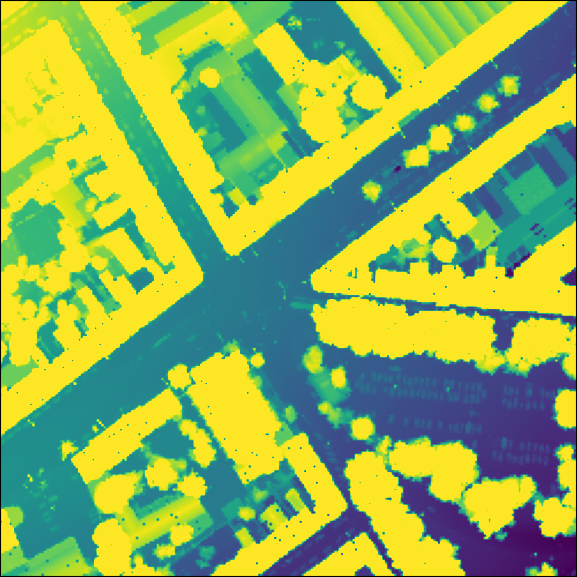}}
        \caption{\ac{DSM}}
        \label{fig:sub2}
    \end{subfigure}
    \begin{subfigure}[b]{0.24\linewidth}
        \centering
        \fbox{\includegraphics[width=\textwidth]{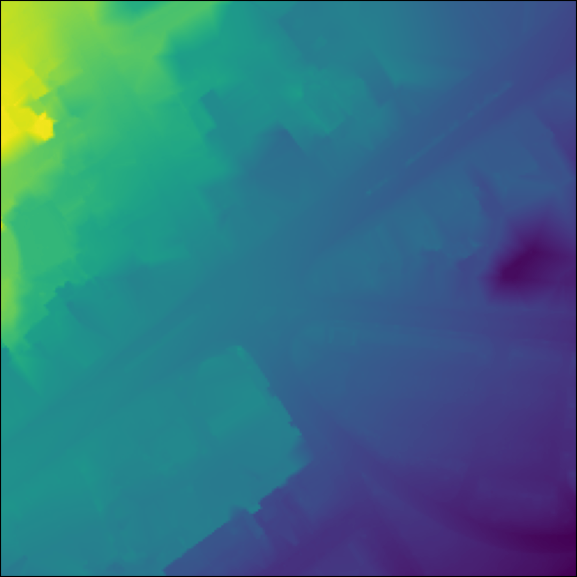}}
        \caption{\ac{DTM}}
        \label{fig:sub3}
    \end{subfigure}
    \begin{subfigure}[b]{0.24\linewidth}
        \centering
        \fbox{\includegraphics[width=\textwidth]{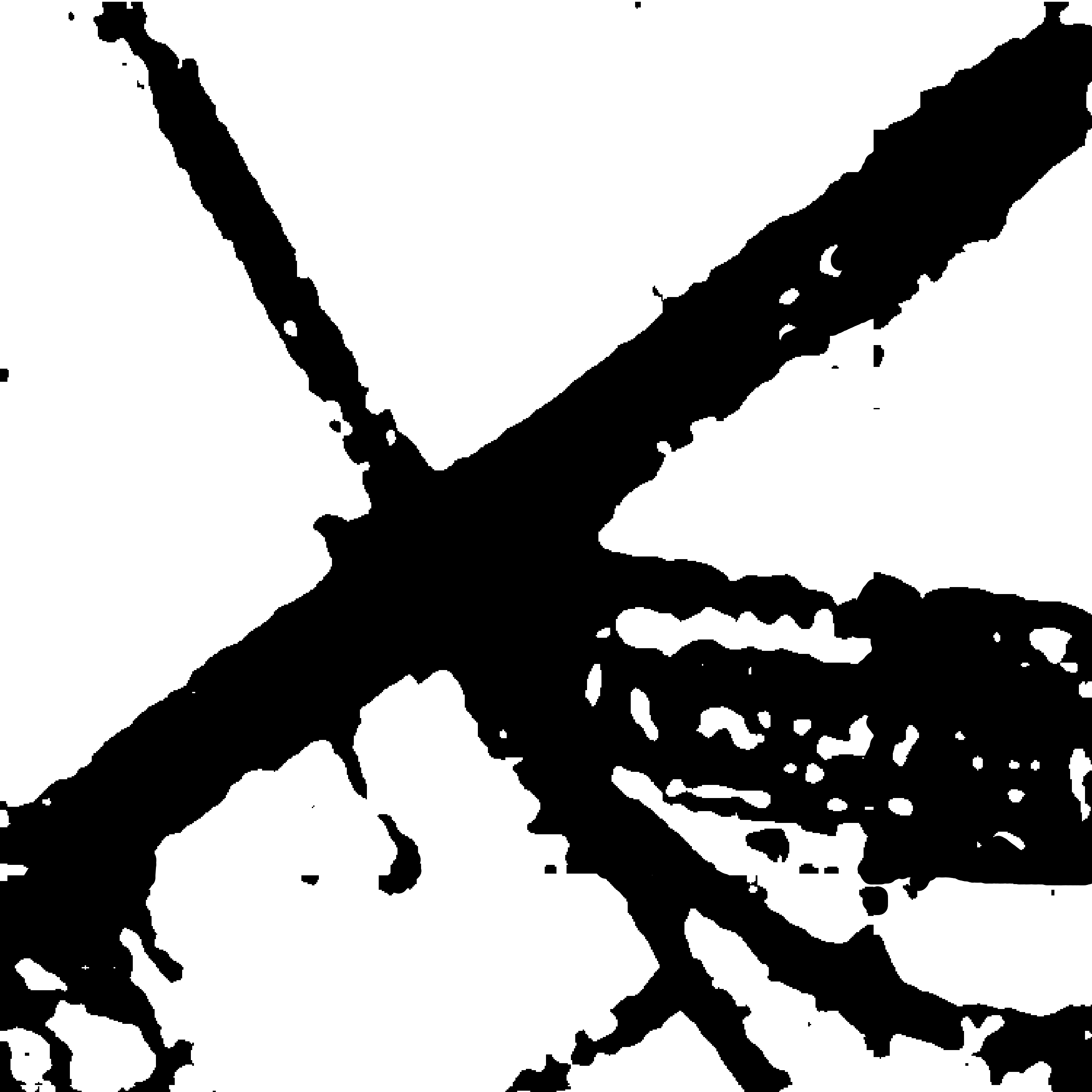}}
        \caption{Road Mask}
        \label{fig:sub4}
    \end{subfigure}
    \caption{\textbf{Sample visualization from \ac{GeoRoad}:} Road intersection in Aachen, Germany\protect\footnotemark. Elevation color coding: blue ($\leq150$m) to yellow ($\geq170$m).}
    \label{fig:example_from_our_dataset}
    \vspace{0.5\baselineskip}
\end{figure}

Our \ac{GeoRoad} dataset\footnote{Dataset available at: \url{https://deepscenario.github.io/FlexRoad/}.} comprises 14,482 million points from Airborne Laser Scanning from Geobasis NRW\footnote{Data from \url{https://www.geoportal.nrw}, under Data License Germany - Zero - Version 2.0.}, including \acp{DSM} and \acp{DTM} with $s_w^{DSM}=s_h^{DSM}=s_w^{DTM}=s_h^{DTM}=1m$ precision and $\pm10cm$ error. High-resolution \acp{DOP} cover 240 locations of $250m\times250m$ tiles across urban (168), sub-urban (23), and rural (49) areas. The dataset features an average elevation difference of 54.7m with roads covering 18\% of the total area. We integrate bridge points from classified airbone \ac{LiDAR} data with \acp{DTM} and extract road masks using DINOv2 \cite{dinov2_2023} to generate $\mathcal{P}^{\road}_{GT}$ and $\mathcal{P}^{\terrain}_{GT}$ from noise-free extended \acp{DTM}. \Cref{fig:example_from_our_dataset} shows an example location with corresponding data representations.

\footnotetext{Data sample: 295500\_5629250, latitude: 50.780040, longitude: 6.100517}

\subsubsection{\acf{DSC}} \label{sec:dsc_dataset}
\ac{DSC}\footnote{Dataset available at: \url{https://app.deepscenario.com/}} comprises five diverse urban environments across Germany and the USA, with photogrammetrically reconstructed meshes of 35K-101K triangles. The scenes include both moderate and steep terrains, with elevation differences from 4.5m to 33.9m. Ground truth generation involves synthetic \ac{LiDAR} sampling, \ac{DSM} and \ac{DTM} generation using \cite{vosselman_2000}, and \ac{DTM} refinement via our \ac{FILTER} algorithm. DINOv2 \cite{dinov2_2023} provides road masks, enhanced through \ac{FILTER}, to extract ground truth $\mathcal{P}^{\terrain}_{GT}$ from \ac{DTM} and $\mathcal{P}^{\road}_{GT}$ from \ac{DSM}.

\subsection{Implementation Details}
We implement our pipeline in PyTorch with a $35 \times 35$ control point grid and cubic basis functions ($p=q=3$). We optimize the loss function for control points $\mathbf{P}_{ij}$ and weights $w_{ij}$ with ADAM optimizer with 0.1 learning rate and 200 iterations with early stopping. We sample surfaces at 1m resolution for roads and $10m$ for terrain. We set the spatial threshold $\theta_{xy} = 10$ such that $ \theta_{xy} > \sqrt{{\scriptstyle {s_w^{DSM}}^2 + {s_h^{DSM}}^2}}$, and $\theta_{z} = 0.5\text{ m}$ encompasses the elevation variations within road clusters. 

\subsection{Evaluation Metrics}
We assess performance on three complementary measures. The $\mathcal{L}_2$ distance quantifies geometric accuracy between the fitted surface and ground truth point clouds ($\mathcal{P}^{\road}_{GT}$, $\mathcal{P}^{\terrain}_{GT}$). While geometric accuracy is crucial, surface smoothness cannot be reliably assessed through comparison with ground truth meshes alone, as these often contain their own reconstruction artifacts and resolution-dependent irregularities. Instead, we evaluate mesh smoothness using the \acf{MAD}:
\begin{align}
MAD = \frac{1}{3n} \sum_{i=1}^n \sum_{j=1}^3 \arccos(| \nabla \mathbf{n}_i \cdot \nabla \mathbf{n}_{ij} |) \, , \label{eq:mad}
\end{align}
where $\mathbf{n}_i$ is the $i^{th}$ face normal and $\mathbf{n}_{ij}$ its $j^{th}$ adjacent face normal. The \ac{MAD} outputs a resolution-independent estimate of local surface $G^2$ continuity as the average angle between adjacent face normals (constrained between 0 and 90 degrees). This intrinsic measure of smoothness is a complement to the $\mathcal{L}_2$ metric, and together they provide both geometric accuracy and surface smoothness necessary for downstream use. We also track mesh complexity through triangle count~$T$, as our goal is to generate lightweight meshes that preserve complex road features while maintaining lower triangle density in less detailed regions.

\subsection{Comparative Analysis}
Our method processes input data from reconstruction methods like COLMAP~\cite{colmap_2016}, OpenSfM~\cite{opensfm_2020}, RoMe~\cite{rome_2023}, EMIE-MAP~\cite{emie-map_2024} or directly from geodata portals, as shown here with the \ac{GeoRoad} (geodata-based) and \ac{DSC} (\ac{SfM}-based) datasets. We compare against two baselines: plane fitting and direct \ac{RGT} from \ac{DSM}, with results in~\Cref{tab:results} and~\Cref{fig:results_plots}.

\input{tables/results.tex}
\input{tables/plots.tex}

On \ac{GeoRoad}, although \ac{RGT} achieves a good accuracy with an $\mathcal{L}_2$ error of 0.415m, it comes at the expense of surface roughness with an \ac{MAD} of 15.55° and high complexity of 124K triangles. Our approach provides a balanced solution with a competitive $\mathcal{L}_2$ error of 0.483m, superior smoothness with an \ac{MAD} of 1.01°, and lower complexity using only 33K triangles. On terrain, \ourmethod reduces the $\mathcal{L}_2$ error of \ac{RGT} by 81\% and significantly improves smoothness, reducing the \ac{MAD} from 22.191° to 3.019°.

On \ac{DSC}, our method achieves state-of-the-art reconstruction quality with an $\mathcal{L}_2$ error of 0.106m for roads and 0.524m for terrain, while providing 94\% better road smoothness than \ac{RGT} with comparable mesh complexity of 40K versus 45K triangles.

Our experiment on both datasets indicates that plane fitting, while generating the smoothest surfaces with minimum complexity, is low in accuracy in all scenarios.

These results highlight the key advantages of our approach: minimal loss of accuracy with a 16\% increase in $\mathcal{L}_2$ for significant smoothness gains showing a 93\% reduction in \ac{MAD} and efficiency with a 73\% reduction in triangles over \ac{RGT} on geodata inputs, with even more impressive results on photogrammetric data. The very large smoothness improvement reflects how effective our approach is at eliminating typical reconstruction noise from non-road object interference while being robust to varying terrain features and input data sources. 

\subsection{Ablation Study}
\input{tables/ablation}

We analyze our method's parameters and components on \ac{GeoRoad} (\Cref{tab:ablation}). Disabling \ac{FILTER} significantly reduces performance, leading to road $\mathcal{L}_2$ error increasing by 111\% and \ac{MAD} rising by 135\%. The absence of \ac{DTM} support causes terrain reconstruction to worsen dramatically by 786\%. When we substitute \ac{NURBS} with \ac{RGT} reconstructions of \ac{DSM} and \ac{DTM} (with road mask), the road \ac{MAD} increases by 1507\% and triangle count rises by 271\%, demonstrating the benefits of our unified parametric approach. Dynamic sampling proves crucial by reducing triangle count by 73\% while maintaining quality.

Parameter analysis reveals several important compromises: using smaller $\theta_z$ values of 0.1m improves accuracy by reducing $\mathcal{L}_2$ by 4\% and enhances smoothness with a 19\% reduction in \ac{MAD}. While these findings suggest potential improvements with $\theta_z = 0.1$, we maintained $\theta_z = 0.5$ in our main experiments for reproducibility, as these ablation studies were conducted after our primary experimental evaluation.

NURBS degree=3 offers the best compromise between accuracy and smoothness compared to lower degrees (which show better $\mathcal{L}_2$ by 18\% but worse roughness) or higher degrees. Grid size significantly impacts quality, with $55 \times 55$ grids yielding 19\% better accuracy but increasing \ac{MAD} by 55\%. Similarly, more detailed sampling (0.5m/5m) improves smoothness by 49\% but doubles the triangle count, while less detailed sampling (2m/20m) reduces triangles by 75\% at the cost of doubling the road \ac{MAD}. These observations are illustrated in \Cref{fig:ablation}.

\input{figures/ablation}

These outcomes demonstrate that surface smoothness and geometric fidelity tend to be conflicting goals, with our \ac{NURBS} approach successfully trading off between them. The \ac{FILTER} performance is shown to be critical for stable surface refinement, while dynamic sampling is responsible for substantial efficiency improvements without great loss in quality. Our pipeline setup is an empirically optimized trade-off of these factors, and can be adjusted via parameter tuning for particular use cases.

%% file: tables/results.tex
\begin{table}[htbp]
\setlength{\tabcolsep}{3pt}
    \vspace{0.5\baselineskip}
    \centering
    \scriptsize
    \caption{\textbf{Quantitative evaluation on \ac{GeoRoad} and \ac{DSC}.}}
    \footnotesize
        \begin{tabular}{l | c | c c | c c | c}
        \toprule
             & \multirow{2}{*}{\textbf{Method}} & \multicolumn{2}{c|}{$\mathbf{\mathcal{L}_2}$ $(m)$ $ \downarrow$} & \multicolumn{2}{c|}{$\textbf{\ac{MAD}}$ $(\text{\textdegree})$ $ \downarrow$} & \multirow{2}{*}{$\mathbf{T} \downarrow$} \\    
             & & \textbf{Road} & \textbf{Terrain} & \textbf{Road} & \textbf{Terrain} & \\    
        \midrule
            \multirow{3}{*}{\adjustbox{angle=90,valign=m}{\acs{GeoRoad}}} 
            & \multirow{1}{*}{Plane} & 2.05 & 2.343 & \textbf{0} & \textbf{0} & \textbf{2}\\
            & \multirow{1}{*}{\ac{RGT}} & \textbf{0.415} & \underline{1.773} & 15.55 & 22.191 & 124K\\
            & \multirow{1}{*}{\ourmethod}~{\scriptsize (Ours)} & \underline{0.483} & \textbf{0.332} & \underline{1.01} & \underline{3.019} & \underline{33K}\\
        \midrule
        \multirow{3}{*}{\adjustbox{angle=90,valign=m}{\ac{DSC}}} 
        & \multirow{1}{*}{Plane} & \underline{1.163} & \underline{2.727} & \textbf{0} & \textbf{0} & \textbf{2} \\
        & \multirow{1}{*}{\ac{RGT}} & 0.195 & 15.411 & 7.185 & 15.78 & 45K \\
        &  \multirow{1}{*}{\ourmethod}~{\scriptsize (Ours)} & \textbf{0.106} & \textbf{0.524} & \underline{0.43} & \underline{1.927} & \underline{40K}\\
            
        \bottomrule
        \end{tabular}
        \label{tab:results}
\end{table}

%% file: tables/plots.tex
\begin{figure}[htbp]
\centering
\vspace{0.5\baselineskip}
\resizebox{1\linewidth}{!}{%
\setlength{\tabcolsep}{1pt}
\begin{tabular}{c c c c}
{\small Plane} & {\small \ac{RGT}} & {\small\ourmethod (Ours)} \\ 
\fbox{\includegraphics[width=0.3\linewidth]{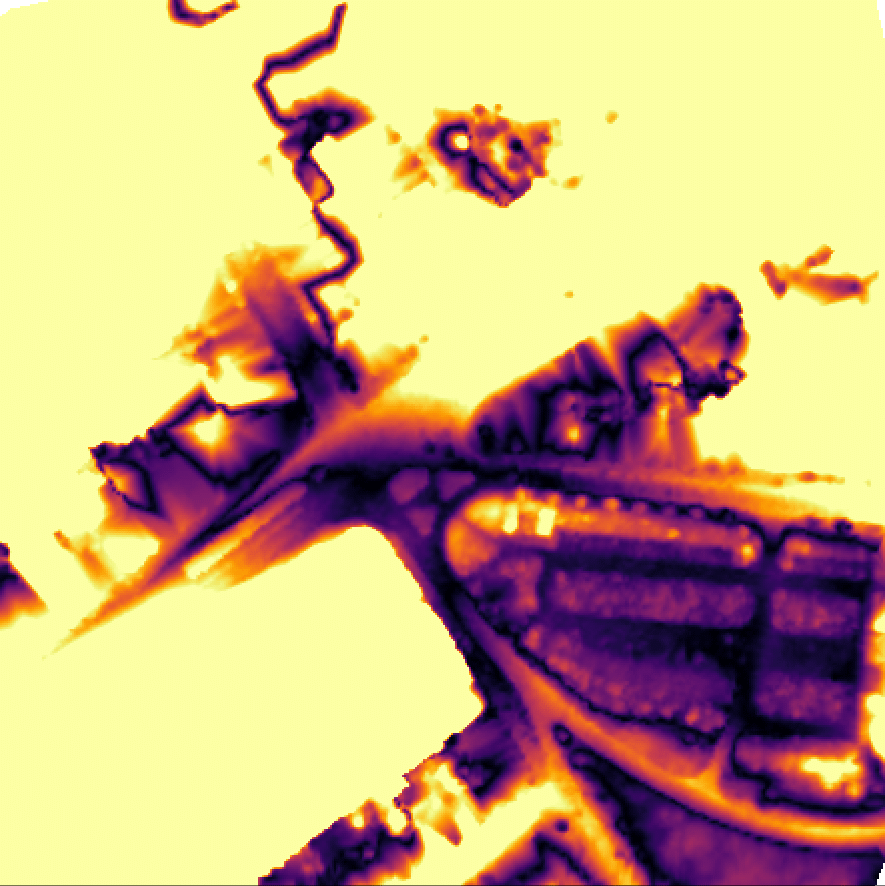}} & \fbox{\includegraphics[width=0.3\linewidth]{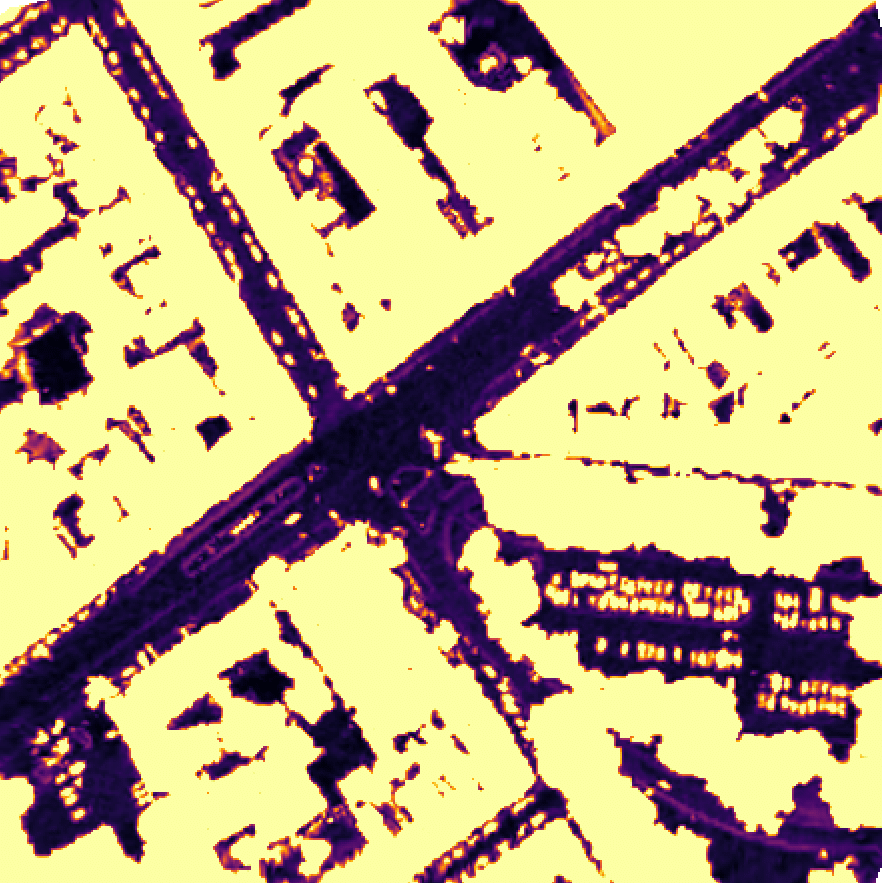}} & \fbox{\includegraphics[width=0.3\linewidth]{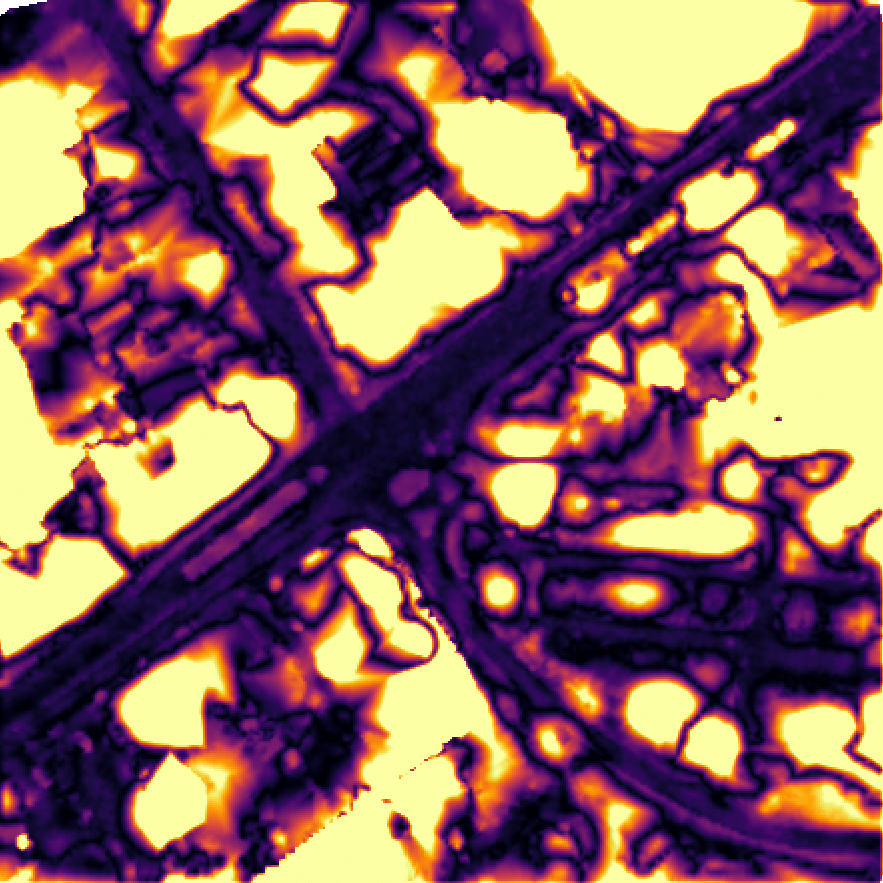}}  & \includegraphics[width=0.06\linewidth]{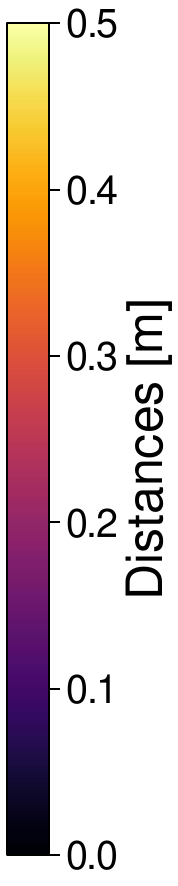}
\\
\fbox{\includegraphics[width=0.3\linewidth]{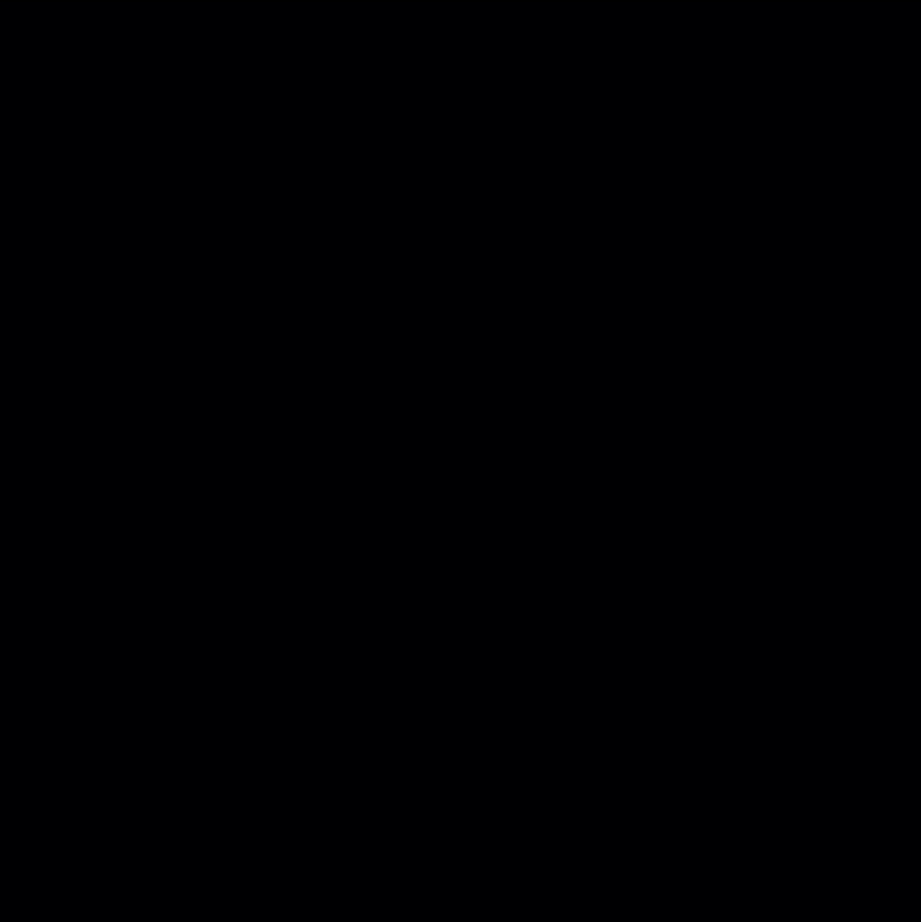}} & \fbox{\includegraphics[width=0.3\linewidth]{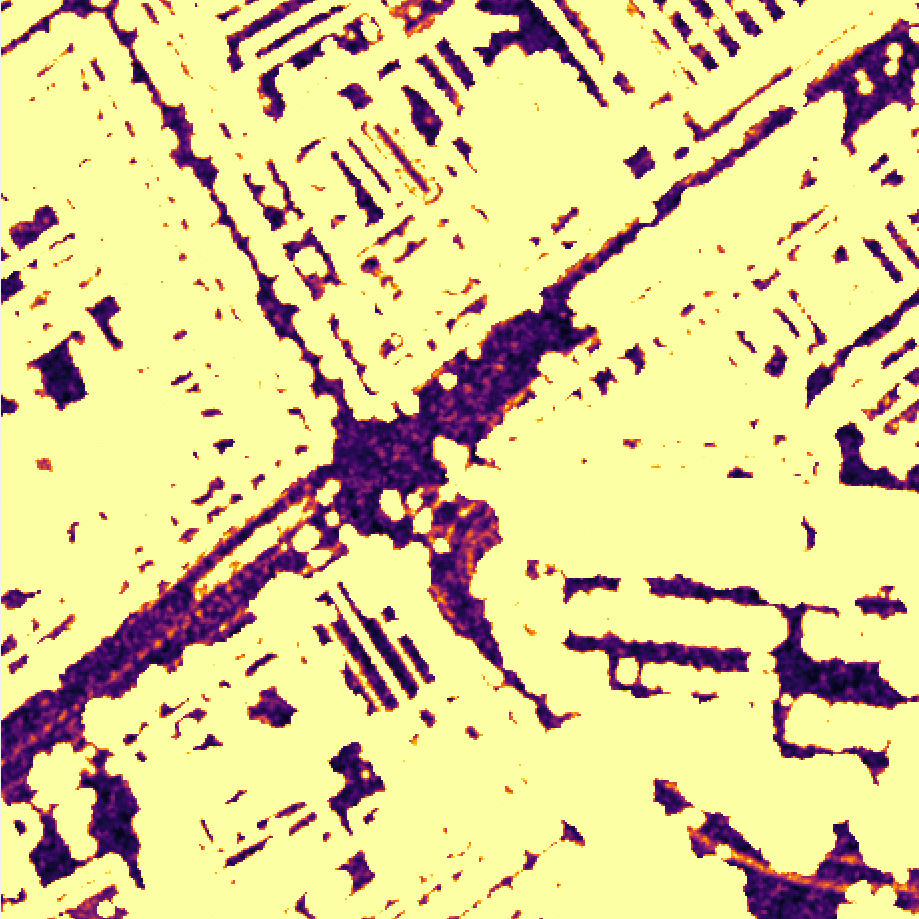}} & \fbox{\includegraphics[width=0.3\linewidth]{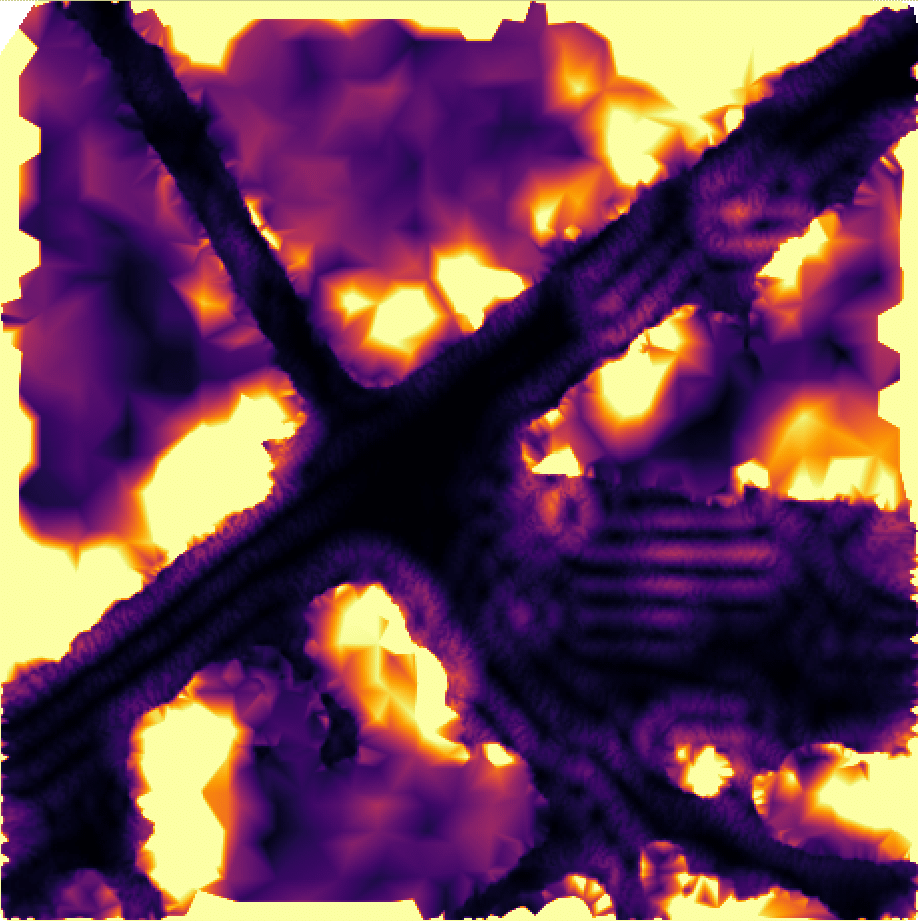}}  & \includegraphics[width=0.06\linewidth]{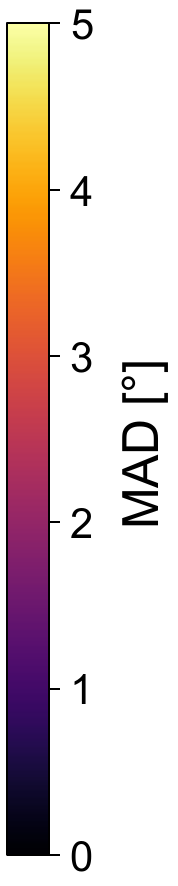}

\end{tabular}
}
\vspace{0.5\baselineskip}
\caption{\textbf{Qualitative results on \ac{GeoRoad}.} Projected $\mathcal{L}_2$ errors (top) and \ac{MAD} smoothness scores (bottom) for sample region 295500\_5629250.}
\label{fig:results_plots}
\end{figure}

%% file: tables/ablation.tex
\begin{table}[htbp]
\setlength{\tabcolsep}{3pt}
    \centering
    \scriptsize
    \caption{\textbf{Ablation study on \ac{GeoRoad}.} Values show relative changes from full pipeline.}
    \label{tab:ablation}
    \footnotesize
    \begin{tabular}{l | c c | c c | c}
        \toprule
        \multirow{2}{*}{\textbf{Variant}} & \multicolumn{2}{c|}{$\mathbf{\mathcal{L}_2} \downarrow$} & \multicolumn{2}{c|}{\textbf{\ac{MAD}} $\downarrow$} & \multirow{2}{*}{$\mathbf{T} \downarrow$} \\
        & \textbf{Road} & \textbf{Terrain} & \textbf{Road} & \textbf{Terrain} & \\
        \midrule
        Full Pipeline & 0.483 &	0.332 &	1.01 &	3.019 &	33,447\\
        \cmidrule(lr){1-6}
        w/o \ac{FILTER} & 1.021 &	1.03 &	2.377 &	5.528 &	33,447 \\
        w/o \ac{DTM} Support & 0.584 & 2.941 & 1.081 & 3.635 & 33,447 \\
        w/o Dynamic Sampling & 0.615 & 1.48 & 0.573 & 0.619 & 122,434 \\
        w/o \ac{NURBS} & 0.472 & 0.207 & 16.23 & 5.7 & 123,953\\
        \cmidrule(lr){1-6}
        %$\theta_{xy}=1.0$m & +X\% & +X\% & +X & +X & X \\
        %$\theta_{xy}=2.0$m & +X\% & +X\% & +X & +X & X \\
        $\theta_z=0.1$m & 0.463 &	0.274 &	0.817 &	2.548 &	33,444 \\
        $\theta_z=1.0$m & 0.489 &	0.331 &	1.006 &	2.943 &	33,447 \\
        $\theta_z=2.5$m & 0.516 &	0.367 &	1.096 &	3.101 &	33,447 \\
        $\theta_z=5.0$m & 0.526 &	0.376 &	1.118 &	3.129 &	33,447 \\
        $\theta_z=10.0$m & 0.549 &	0.381 &	1.142 &	3.142 &	33,447 \\
        \cmidrule(lr){1-6}
        Degree: p=q=2 & 0.395 &	0.317 &	1.118 &	3.382 &	33,459 \\
        Degree: p=q=5 & 0.484 &	0.351 &	0.943 &	2.585 &	33,434 \\
        Degree: p=q=10 &0.491 &	0.449 &	0.853 &	2.251 &	33,406 \\
        \cmidrule(lr){1-6}
        Grid: $15\times15$ & 0.553&	0.616&	0.635&	2.114&	33,504 \\
        Grid: $25\times25$ & 0.502 &	0.411 &	0.79 &	2.558 &	33,614 \\
        Grid: $45\times45$ & 0.474 &	0.301 &	1.302 &	3.571 &	33,591 \\
        Grid: $55\times55$ & 0.392 &	0.282 &	1.568 &	3.964 &	33,461 \\
        \cmidrule(lr){1-6}
        Sampling: 0.5m/5m & 0.483 &	0.326 &	0.512 &	1.521 &	134,113 \\
        Sampling: 2m/20m & 0.482 &	0.393 &	1.985 &	5.145 &	8,228 \\
        \bottomrule
    \end{tabular}
    \vspace{1\baselineskip}
\end{table}

%% file: figures/ablation.tex
\begin{figure}[!ht]
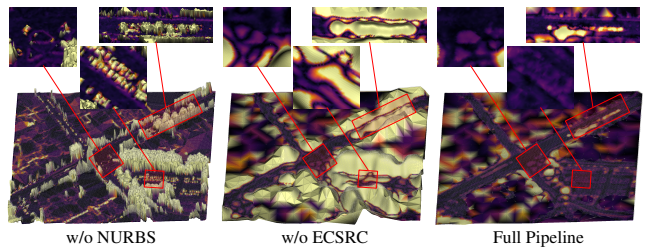

\centering
\vspace{0.5\baselineskip}
\resizebox{1\linewidth}{!}{%
\setlength{\tabcolsep}{1pt}
\begin{tabular}{c c c}
\includegraphics[width=0.33\textwidth]{figures/ablation_no_nurbs.pdf} & \includegraphics[width=0.33\textwidth]{figures/ablation_no_filter.pdf} & \includegraphics[width=0.33\textwidth]{figures/ablation_full.pdf} \\
{\Large w/o \ac{NURBS}} & {\Large w/o \ac{FILTER}} & {\Large Full Pipeline}
\end{tabular}
}
\vspace{0.5\baselineskip}
\caption{\textbf{Road reconstruction comparison} on \ac{GeoRoad} region 295500\_5629250. The vertex colors indicate $\mathcal{L}_2$ errors using the same color scheme as \Cref{fig:results_plots}.}
\label{fig:ablation}
\end{figure}

%% file: chapters/5_conclusion.tex
\section{Conclusion}
\label{sec:conclusion}
In this paper, we presented \ourmethod, a novel road surface refinement method that employs \ac{NURBS} surface fitting and our geometric filtering \ac{FILTER} algorithm to preserve the natural smoothness of road surfaces at high accuracy. On our newly introduced \ac{GeoRoad} dataset, our method improved road accuracy by 76\% compared to plane fitting with 93\% less surface roughness than \ac{RGT}. These improvements were achieved with 73\% fewer triangles through dynamic sampling. The accuracy of the technique was also evaluated on the photogrammetry reconstruction-based \ac{DSC} dataset, where it demonstrated high accuracy and smoothness over diverse urban scenes. Ablation studies confirmed the contribution of each element in achieving this trade-off between accuracy and smoothness. This work provided a solid basis for surface refinement and opened up future research opportunities. Future work will focus on road intersections with complex structures, multi-level road structures, and real-time mobile mapping applications, as well as applying refinement steps to extract detailed road features such as cracks and road surface degradation.